\documentclass{article}
\usepackage[utf8]{inputenc}
\usepackage{authblk}
\usepackage[numbers,sort&compress]{natbib}
\usepackage{graphicx}
\usepackage{amsmath}
\title{Solution of Definite Integrals using Functional Link Artificial Neural Networks}
\author[1]{Satyasaran Changdar}
\author[1]{Snehangshu Bhattacharjee}
\affil[1]{Department of Information Technology, Institute of Engineering and Management, Salt Lake, Kolkata}

\date{}

\providecommand{\keywords}[1]
{
  \small	
  \textbf{\textit{Keywords---}} #1
}

\begin{document}

\maketitle

\begin{abstract}
This paper discusses a new method to solve definite integrals using feedforward artificial neural networks. The objective is to build a neural network that would be a novel alternative to pre-established numerical methods and with the help of a learning algorithm, be able to solve definite integrals, by minimising a well constructed error function. The proposed algorithm, with respect to existing numerical methods, is effective and precise and well-suited for purposes which require integration of higher order polynomials. The observations have been recorded and illustrated in tabular and graphical form.\\
\end{abstract}

\section{Introduction}
Through centuries, men have been captivated by the power of the human brain. Our very ability to capture, comprehend, analyse and draw conclusions from seemingly innocent pieces of information, is a marvel which deserves admiration. Only recently has science been able to gauge intelligence and its parameters, and yet it is far from fathoming the intricate machinery of sense and logic completely.\\
With several scientific attempts being made to study activities and patterns of the brain since the nineteenth century, it was only in 1943 that McCulloch and Pitts introduced a simple model of a neural network \citep{mcculloch}, which had weighted inputs summed up to give the output. With the advent of computers in the 1950s, it was finally possible to implement the theories.
The concept of a "thinking machine" has been much argued over, and the Dartmouth Project in 1956 followed by Frank Rosenblatt's work in development of a Perceptron \citep{rosenblatt} was a major impetus to the idea. Mark I Perceptron was built in the year 1958, at Cornell University, with its main function being character recognition. Soon after, Bernard Widrow and Marcian Hoff of Stanford successfully launched neural network models ADALINE and MADALINE \citep{madaline} for real-world purposes.\\
In 1982, John Hopfield of Caltech, with his charismatic paper presented to the National Academy of Sciences \citep{hopfield}, started rekindling interest in what was, by far, an overlooked field in science. The US-Japan Joint Conference on Cooperative/Competitive Neural Networks propelled efforts considerably and by 1985, the Neural Networks for Computing conference was held, which was to become an annual affair. Also, IEEE's first International Conference on Neural Networks became a success in 1987. The last decade of the twentieth century was crucial in bringing the study of neural networks to the fore with many sectors in innovation and research recognising and targeting it. \\
ANN is being extensively used as a tool in various disciplines for applications which include modelling complex mathematical structures and obtaining solutions to mathematical problems. Mathematical models, with a framework of ANN, can provide sustainable and effective solutions in commercially applicable fields \citep{app1} \citep{satya} \citep{satya1} \citep{satya3}. Functions such as those discussed below are frequently encountered in real-life problems of engineering. Calculus is extensively used in manufacturing and other core sectors for applications like improving architecture and determining the infrastructure of bridges. It plays an important role in applications that deal with massive data and require precision and calculus equations to make predictions like in meteorology and in the finance sector. Even in the case of search engines which demand a number of variables to be taken into account, calculus comes into play. The next logical step would be incorporation of neural networks in practical applications in the near future.\\
In the present paper, we attempt to discuss a way of solving Definite Integrals by applying a direct, clinical and robust approach using the effectiveness of neural networks instead of more cumbersome numerical methods involving analytical calculations.

\section{Motivation}
Determining an Indefinite Integral refers to the process of determining a function from its derivative \emph{f(x)} \citep{calc1}. The functions that could possibly have \emph{f(x)} as a derivative are the so-called antiderivatives of \emph{f(x)},and the formula that leads to them is called the indefinite integral of \emph{f(x)}.\\
Mathematically:
\begin{equation}
\int{f(x)dx}=g(x)+C
\end{equation}\\
In this equation, \emph{g(x)} is one of the antiderivatives of \emph{f(x)}, and as the constant \emph C changes, the equation gives a family of antiderivatives.\\
Let \emph{g(x)} be the antiderivative of the continuous function \emph{f(x)} defined on \emph{[a,b]}. Then the definite integral of \emph{f(x)} over \emph{[a,b]} is denoted by
\begin{equation} 
\int_{a}^{b}{f(x)dx}  = g(b) - g(a)
\label{eq:2}
\end{equation}

In the process of solving the above equation, we come across a number of problems:\\
1. The integrand \emph{f(x)} may be known only at certain points.\\
2. A formula for the integrand \emph{f(x)} may be known, but it may be difficult or impossible to find an antiderivative, that is, an elementary function.\\
3. It may be possible to find an antiderivative symbolically, but it may be easier to compute a numerical approximation than to compute the antiderivative.\\
In the proposed method, we have discussed a solution in accordance to the above points \emph{(2)} and \emph{(3)}.\\
Usually, in cases where analytical solutions may not be arrived at easily, we rely on numerical methods for solving the equation. This again leads to specialised methods, specific to the purpose. A more appropriate approach would be to design a solution that aptly fits in the places of these varied methods and gives a decently precise and accurate solution. This leads us to consider Artificial Neural Networks as an alternative \citep{mall0} \citep{mall1} \citep{mall2}.
The efficiency of Artificial Neural Networks is due to its nonlinear learning capability \citep{annbook}. This powerful tool finds application in functional approximation and optimisation. Due to application of FLANN (Functional Link Artificial Neural Network) \citep{mall3} \citep{sahoo1}, we no longer need hidden layers. This makes the learning algorithm simpler. The FLANN architecture effectively expands the input vector giving more flexibility and decision-making capability to the network, due to a more expansive input space.\\

\section{Mathematical modelling}
We have already established that the definite integral of a function can be calculated as shown in equation \ref{eq:2} 
\begin{equation}
\int_{a}^{b}{f(x)dx}  = g(b) - g(a)
\end{equation}
In the proposed model we would replace the \emph{g(b)} and \emph{g(a)} by \emph{N(b)} and \emph{N(a)} respectively.
Here, \emph{N(x)} represents a function of \emph{x} which is obtained as the output of the neural network.
Therefore our former equation reduces to 
\begin{equation}
\int_{a}^{b}{f(x)dx}  = N(b) - N(a)
\label{eq:3}
\end{equation}\\
\subsection{Model of the Neural Network}
Output of the Neural Network
\begin{equation}
N(x)= w_1\Phi_1(x) + w_2\Phi_2(x) + w_3\Phi_3(x) +......+ w_n\Phi_n(x)
\label{eq:4}
\end{equation}\\
In general, a model of a neural network comprises simple computational units (nodes) which may be input nodes, output nodes or hidden nodes. The units, with appropriate interconnections, is represented by a graph with the nodes as the vertices and each edge between two neurons is labelled with a synaptic weight. The network may or may not have multiple layers, depending on the complexity of problems. As layers and number of neurons increase, the training process becomes more complicated and requires intense computation. An alternative to this cumbersome approach is the FLANN architecture \citep{mall3}\citep{sknanda1}\citep{sknanda2}.
In our present study, we have employed the classification method of Functional Link Artificial Neural Network (FLANN). FLANN architecture uses a single layer feed-forward network, which can be considered as an alternative to multi-layer neural network for solving problems. The drawback with multi-layer neural network is its linear nature which becomes a hindrance for mapping complex non-linear problems. Since the practical realm of data deals mostly with non-linearity, therefore, choosing multi-layer networks for this is not wise. Yet again, employing a multi-layer neural network for this purpose further complicates the task. But in FLANN architecture, the input vector is functionally expanded to overcome such problems, such that even though it is a single layer neural network, it is well-suited to the task. The model which has been used is shown in Fig.1.
\begin{figure}[!ht]
\centering
\includegraphics[scale=0.25]{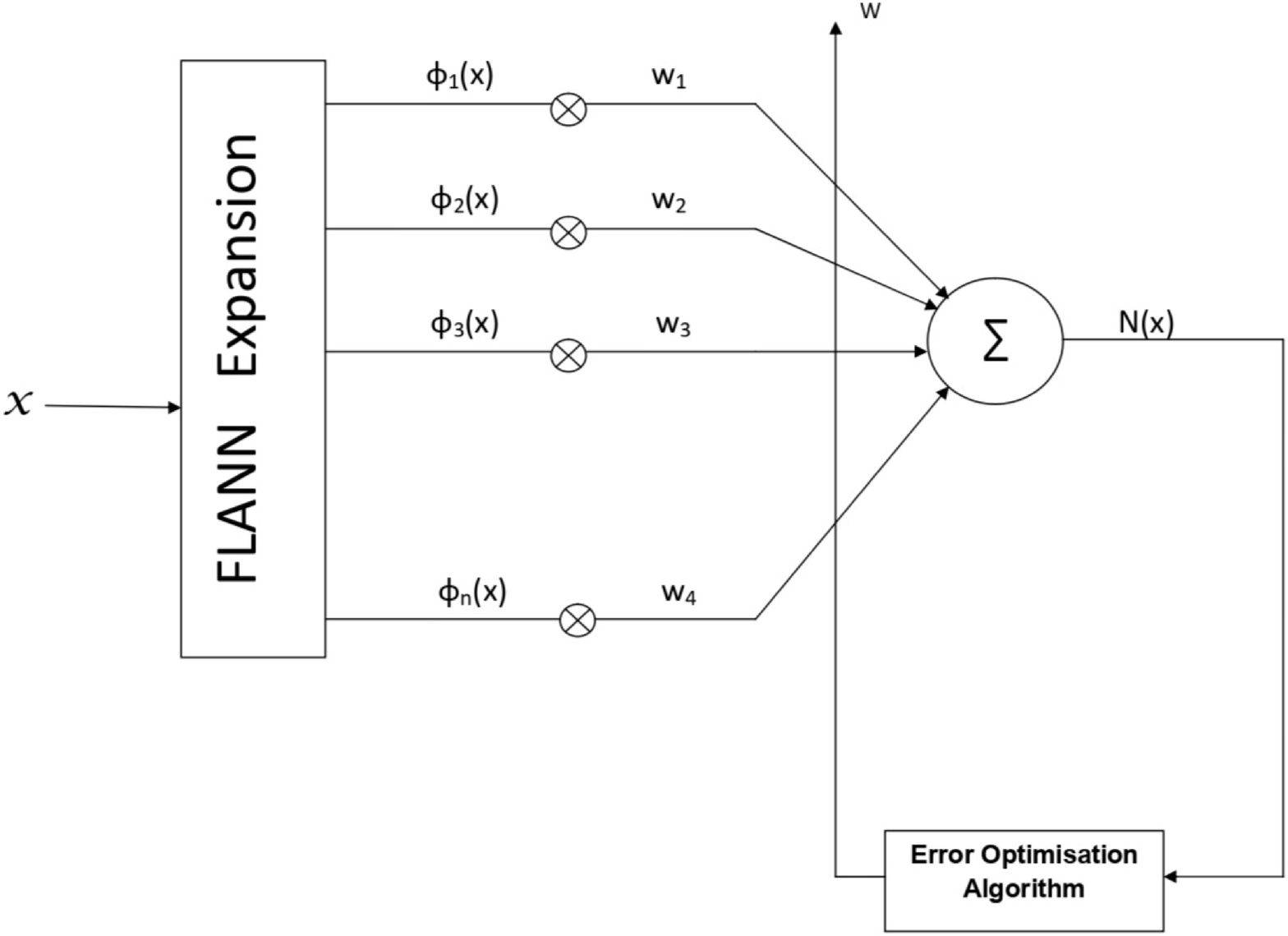}
\caption{Schematic diagram of the neural network model}
\end{figure}
\\
If the input vector is \emph{x}, then after expansion, the input pattern is a set of elements :
\begin{equation}
\Phi_1(x), \Phi_2(x),.....\Phi_n(x)
\end{equation}
which are appropriate functions that will serve in the approximation process. The set of functions for expansion may be set in accordance to the complexity of the problem and the restraints posed on it.

\subsection{Modelling the Error Function}
As discussed above in equation \ref{eq:3} 
\begin{equation}
\int_{a}^{b}{f(x)dx}  = N(b) - N(a)
\end{equation}
Our objective is to build a function \emph{N(x)} using the neural network \citep{lagaris1} \citep{lagaris2} \citep{yadav} \citep{yadav1} which will approximate the function \emph{g(x)} upto a desired precision, so that this function \emph{N(x)} can be later on employed to compute the values of the integral at \emph{a} and \emph{b}.  Therefore,
\begin{equation}
g(x) \approx N(x)
\end{equation}
In the ideal scenario, \emph{g(x)} equals \emph{N(x)}
\begin{equation}
g(x)-N(x)=0
\label{eq:6}
\end{equation}
Differentiating equation \ref{eq:6} with respect to \emph{x} gives,\nolinebreak
\begin{equation}
g'(x)-N'(x)=0
\end{equation}
or,\nolinebreak
\begin{equation}
f(x)-N'(x)=0
\label{eq:8}
\end{equation}
where,\nolinebreak
\begin{equation}
N(x)=w_1\Phi_1(x) + w_2\Phi_2(x) + w_3\Phi_3(x) +......+ w_n\Phi_n(x)
\end{equation}
\begin{equation}
N'(x)=w_1\Phi_1'(x) + w_2\Phi_2'(x) + w_3\Phi_3'(x) +......+ w_n\Phi_n'(x)
\label{eq:9}
\end{equation}\\
Therefore equation \ref{eq:8} becomes
\begin{equation}
f(x)-(w_1\Phi_1'(x) + w_2\Phi_2'(x) + w_3\Phi_3'(x) +......+ w_n\Phi_n'(x))=0
\end{equation}
The error function can now be summarised as:
\begin{equation}
E(x) = \frac{1}{2}{\sum_{i=1}^{k}{{(f(x_i)-N'(x_i))}}^2}
\label{eq:10}
\end{equation}
where $x_i$ refers to each training sample point; \emph{k} points belonging to the interval \emph{(a,b)} have been considered.

\subsection{Minimisation of Error}
In this study, we have already constructed an error function, minimisation thereof would lead to an appropriate solution to the given definite integral. The error function uses a parameter which corresponds to the result derived from the neural network. This parameter is modified with every modification of the network (modification here corresponds to updation of the weights in the network).
The error optimisation algorithm used is gradient descent algorithm \citep{gradient1} \citep{gradient2}.
This algorithm uses first-order iterative optimisation to find the minimum of a function. The algorithm is designed to proceed in steps proportional to the gradient of the function, in the negative direction of the current point. 
In the gradient descent algorithm, initially, the gradient of the function is to be calculated. This is done by differentiating the error function with respect to each of the weights one by one. These are then used to update the weights. The process leads to reduction of the error gradually until we reach a desired precision.\\ \\
Differentiating \ref{eq:10} with respect to each weight:
\begin{equation}
\frac{dE}{dw_1} = -\sum_{i=1}^{k}{e_i \Phi_1'(x_i)}
\end{equation}
\begin{equation} 
\frac{dE}{dw_2} = -\sum_{i=1}^{k}{e_i \Phi_2'(x_i)}
\end{equation}
\begin{equation}
\frac{dE}{dw_3} = -\sum_{i=1}^{k}{e_i \Phi_3'(x_i)}
\end{equation}
and likewise, till
\begin{equation}
\frac{dE}{dw_n} = -\sum_{i=1}^{k}{e_i \Phi_n'(x_i)}
\end{equation}
\\
where, \begin{equation}\emph{$e_i$}=f(x_i)-N'(x_i)\end{equation}
The weights are then updated as:
\begin{equation}
w_1^+ = w_1 - \eta \frac{dE}{dw_1}
\end{equation}
\begin{equation}
w_2^+ = w_2 - \eta \frac{dE}{dw_2}
\end{equation}
\begin{equation}
w_3^+ = w_3 - \eta \frac{dE}{dw_3}
\end{equation}
and likewise, till
\begin{equation}
w_n^+ = w_n - \eta \frac{dE}{dw_n}
\end{equation}\\
where, $\eta$ is the \emph{learning rate}.

\section{Results and observations}
Let us work out an example using the method discussed above. Consider the function,\\
\begin{equation} f(x)=\sqrt{1+(x^2)}
\label{eq:12}
\end{equation}\\
We would like to integrate the function from 0 to 2, i.e
\begin{equation}
\int_{0}^{2}{\sqrt{1+(x^2)}dx}
\end{equation}\\
Using Maclaurin Series, a function \emph{g(x)} can be expanded as,
\begin{equation}
g(x)=\sum_{n=0}^{\infty}{g^{(n)}(0)\frac{x^n}{n!}}
\end{equation}
where $g^{(n)}$ is the $n^{\text{th}}$ derivative of \emph{g(x)}.\\
Expanding the concerned function using Maclaurin's Series,
\begin{equation}
\sqrt{(1+(x^2))}=1+\frac{x^2}{2}-\frac{x^4}{8}+\frac{x^6}{16}-.....
\label{eq:13}
\end{equation}\\

Approximating equation \ref{eq:13} using the neural network would require a good choice of the functional links $\phi_1, \phi_2, \phi_3, .... \phi_n$\\
In accordance to the Maclaurin Series expansion of \emph{f(x)}, which is a polynomial function, we choose\\
\begin{equation}
\phi_1(x)=x
\end{equation}
\begin{equation}
\phi_2(x)=x^2
\end{equation}
\begin{equation}
\phi_3(x)=x^3
\end{equation}
and likewise, till
\begin{equation} 
\phi_n(x)=x^n
\end{equation}\\
Therefore, equation \ref{eq:4} can be reconstructed as\\
\begin{equation}
N(x)= w_1x + w_2x^2 + w_3x^3 +......+ w_nx^n
\end{equation}\\
Similarly, equation \ref{eq:9} becomes
\begin{equation}
N'(x)=w_1 + 2w_2x + 3w_3x^2 +......+ nw_nx^{n-1}
\end{equation}\\
The error function for the problem under consideration, becomes\\
\begin{equation}
E(x) = \frac{1}{2}{\sum_{i=1}^{k}{(f(x)-(w_1 + 2w_2x + 3w_3x^2 +..+ nw_nx^{n-1}))^2}}
\end{equation}\\
The objective is to optimise this error function for which we initialise the network with random weights, a suitable learning rate $\eta$ is chosen, and with every iteration, the weights are updated using the gradient descent algorithm, as follows:
\begin{equation}
w_1^+ = w_1 - \eta \frac{dE}{dw_1}
\end{equation}
\begin{equation}
w_1^+ = w_1 - \eta \sum_{i=1}^{k}{-e_i}
\end{equation}
\begin{equation}
w_2^+ = w_2 - \eta \frac{dE}{dw_2}
\end{equation}
\begin{equation}
w_2^+ = w_2 - \eta \sum_{i=1}^{k}{-e_i(2x_i)}
\end{equation}
\begin{equation}
w_3^+ = w_3 - \eta \frac{dE}{dw_3}
\end{equation}
\begin{equation}
w_3^+ = w_3 - \eta \sum_{i=1}^{k}{-e_i(3{x_i}^2)}
\end{equation}
and likewise, till
\begin{equation}
w_n^+ = w_n - \eta \frac{dE}{dw_n}
\end{equation}
\begin{equation}
w_n^+ = w_n - \eta \sum_{i=1}^{k}{-e_i(n{x_i}^{n-1})}
\end{equation}

\begin{figure}[ht]
\centering
\includegraphics[scale=0.5]{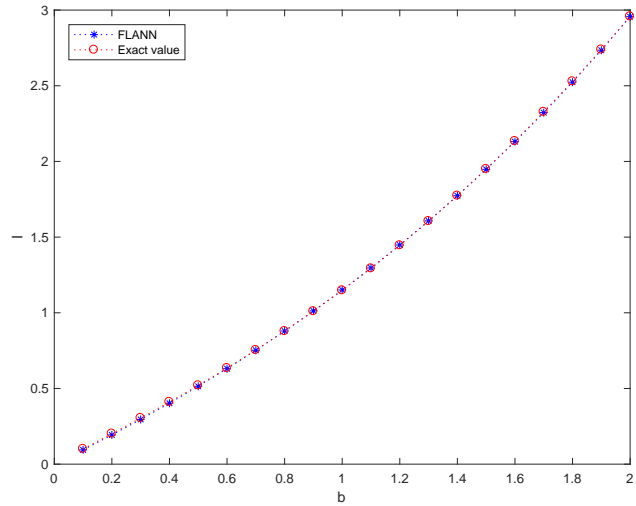}
\caption{Graph comparing exact value with observed value using FLANN for $I=\int_{0}^{b}{f(x)dx}$ where $f(x)=\sqrt{1+x^2}$,  b$\in$(0,2]}
\label{fig.2}
\end{figure}

The above procedure is repeated till the error is minimised to a desired precision.
Finally, the weights are locked at the most favourable magnitudes (where the error function is minimised). These weights, when now incorporated in the network (i.e. $N(x)$ which has been approximated to $g(x)$, the assumed anti-derivative of $f(x)$), provides us the result of the integral in any interval within \emph{[0,2]}, i.e, the domain in which the network has been trained.
The results obtained from this neural network versus the exact values calculated analytically, are illustrated in Fig.\ref{fig.2}
\\
\\
\\

\begin{figure}[!ht]
\centering
\includegraphics[scale=0.41]{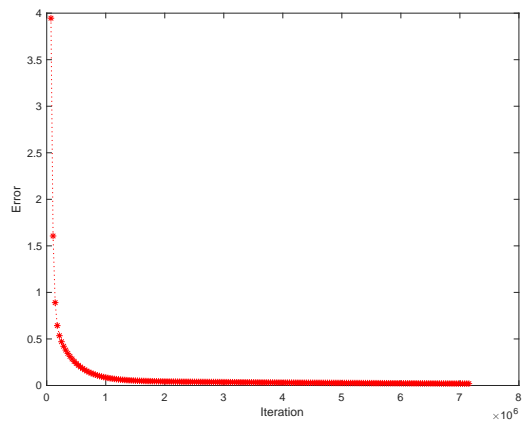}
\caption{Error vs. Iteration while integrating $\sqrt{1+x^2}$}
\label{fig.3}
\end{figure}

On analysing the change in error with the number of iterations, we obtain Fig.\ref{fig.3}

\begin{figure}[!ht]
\includegraphics[scale=0.5]{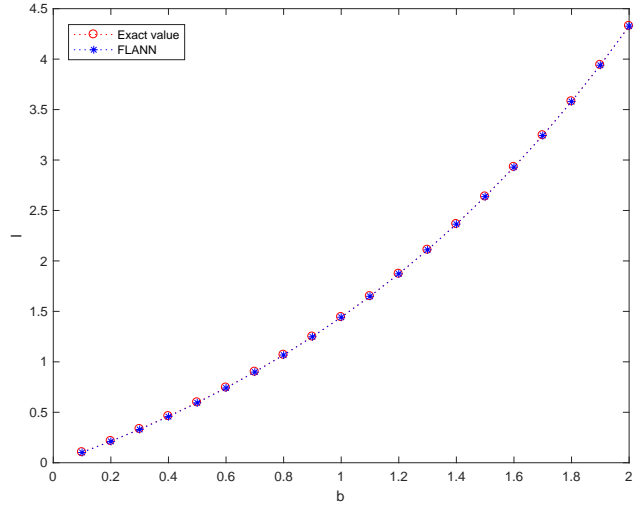}
\centering
\caption{Graph comparing exact value with observed value using FLANN for $I=\int_{0}^{b}{f(x)dx}$ where $f(x)=2^x$,  b$\in$(0,2]}
\label{fig.4}
\end{figure}

\begin{figure}[!ht]
\centering
\includegraphics[scale=0.6]{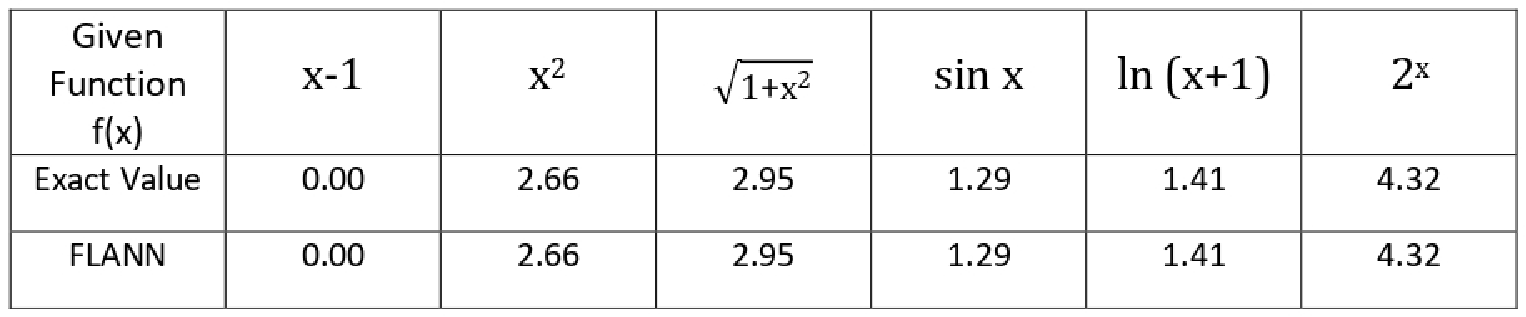}
\caption{Comparison of a few functions solved analytically and using the discussed model}
\label{fig.5}
\end{figure}

Similarly, the proposed algorithm is applied to integrate another function:
\begin{equation}
\int_{0}^{2}{{2^x}dx}
\end{equation}\\\\
The Fig.\ref{fig.4} is obtained in this case.
\\
\\
In order to illustrate the feasibility and efficiency of the proposed method, a few examples are selected, the method is applied to them, and the results are recorded in Fig.\ref{fig.5}. The given functions are integrated over [0,2].
\\
\\

\section{Examples for comparison with other methods}
Conventional numerical methods, when implemented in calculating integrals, are not always sufficient or accurate in approximation\citep{num2}. For instance, the Trapezoidal rule gives an error:
\begin{equation*}
E^T_n(f)\equiv\int^b_a{f(x)dx}-T_n(f)\equiv-\frac{h^2(b-a)}{12}f^2(c_n)
\end{equation*}
for some $c_n$ in the interval (a,b).\\
\\
As is evident, the error in Trapezoidal rule increases in the following scenarios:\\
1. If $h$ increases, where $h$ is the width of each subinterval between $[a,b]$\\
2. If $(b-a)$ is a large value\\
3. If the function $f(x)$ is a polynomial of a high degree; which causes $f^2(c_n)$ to yield a large value.\\

\begin{figure}[!ht]
\centering
\includegraphics[scale=0.45]{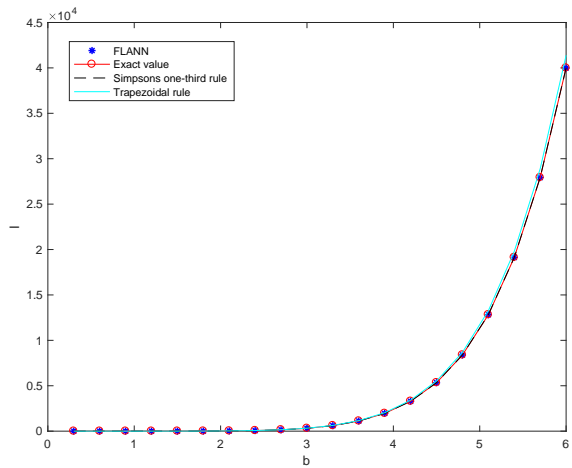}
\caption{Graph comparing the results of the integral $I=\int_{0}^{b}{f(x)dx}$ where $f(x)=x^6$,  b$\in$(0,6] calculated using FLANN, Trapezoidal method, Simpson's rule and the same calculated analytically}
\label{fig.6}
\end{figure}

\begin{figure}[!ht]
\centering
\includegraphics[scale=0.45]{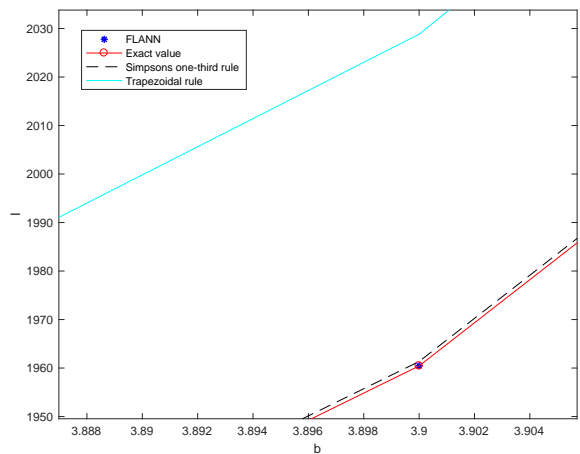}
\caption{This graph is a magnified version of Fig.6, which clearly shows the precision attained by the proposed FLANN algorithm when compared to other techniques. Note that the exact value has completely coincided with the calculated value using the FLANN technique.}
\label{fig.7}
\end{figure}

Similarly, Simpson's One-third rule gives an error:\\
\begin{equation*}
E^S(f)\equiv\int^b_a{f(x)dx}-S(f)\equiv-\frac{1}{180}(b-a)h^4f^4(c_n)
\end{equation*}
where $c_n$ lies in the interval (a,b).\\
\\
The error, in this case increases for the following:\\
1. If $h$ increases, where $h$ is the width of each subinterval between $[a,b]$\\
2. If $(b-a)$ is a large value\\
3. If the function $f(x)$ is a polynomial of a very high degree, which causes $f^4(c_n)$ to yield a large value.\\
\\
In the following example, we have considered the function\\
\begin{equation}
\int^b_a{f(x)dx}=\int^6_0{x^6dx}
\label{eq:16}
\end{equation}
Since the polynomial considered is of a high degree, therefore $f^2(c_n)$ and $f^4(c_n)$ yields a considerably larger value where $c_n$ lies in the interval $(0,6)$. In cases where $f(x)$ is a higher degree polynomial and $(b-a)$ is also large, both the Trapezoidal and Simpson's Rule would fail to come up with precise results.\\
\\
To show the promising results of the proposed FLANN architecture, we have compared the results of the equation \ref{eq:16} with those obtained from FLANN, Trapezoidal method, Simpson's rule and the same calculated analytically. While training the neural network for the concerned problem, we have considered 10 training points \emph{(i.e. k=10)} belonging to the interval (a,b). Similarly, using \emph{k} points in Trapezoidal and Simpson's Rule implies that the interval $[a,b]$ gets divided into \emph{k+1} sub-intervals, thus the width \emph{h} of each subinterval becomes  $\frac{b-a}{k+1}$. These results have been shown in Fig.\ref{fig.6} and Fig.\ref{fig.7}.\\

\section{Practical Applications}
While solving real life problems we come across various situations where solving a definite integral analytically becomes difficult or seems impossible. To tackle situations like these we have been using numerical integration techniques. One such definite integral which is difficult to solve analytically is the elliptic integral. Mathematics defines an elliptic integral as any function f(x) which can be expressed in the form
\begin{equation}
f(x) = \int_{c}^{x}{R(t,\sqrt{P(t)}) dt}
\end{equation}
where R is a rational function of its two arguments, P is a polynomial of degree 3 or 4 with no repeated roots, and c is a constant. In general, integrals in this form cannot be expressed in terms of elementary functions easily. This is where numerical integration techniques proves to be of immense help and can often give precise results. Integrals like elliptic integrals which are often difficult to solve have huge applications in astrophysics and various engineering domains.

\section{Conclusion}
The main objective of the paper has been to establish a new method using neural networks, which is applicable in solving definite integrals in a more efficient and novel approach as compared to existing methods. We construct a neural network model based on FLANN architecture, which is used to approximate the integrand by training the neural network weights. The solved examples, compared with other existing methods, show that the algorithm is effective and feasible in all cases.\\
The proposed algorithm has the following advantages:\\
1. The FLANN architecture used successfully incorporates non-linearity in the model, which is a requirement in most practical scenarios.\\
2. The neural network, when trained over the interval [a,b], can easily estimate the values in another interval [$a_1$,$b_1$] where a $\leq$ $a_1$,$b_1$ $\leq$ b.\\
3. Even when computing integrals of higher order polynomials, the method evaluates to highly precise levels.\\
\\
The successful implementation of the algorithm can depend on certain improvisations, as follows:\\
1. Change of optimisation method.\\
2. Better choice of the functional links.\\
3. Use of high-speed and more reliable processors.\\

\bibliographystyle{plain}
\bibliography{references}
\end{document}